\documentclass[11pt]{article}

\usepackage[preprint]{acl}

\usepackage{times}
\usepackage{latexsym}
\usepackage{arydshln}
\usepackage{booktabs}
\usepackage{lipsum}

\usepackage[T1]{fontenc}

\usepackage[utf8]{inputenc}

\usepackage{microtype}

\usepackage{inconsolata}

\usepackage{graphicx}
\usepackage{enumitem}
\usepackage{amsmath}
\usepackage{amssymb}
\usepackage{booktabs}
\usepackage{subcaption}
\usepackage{multirow}
\usepackage{comment}

\usepackage[table]{xcolor}

\usepackage{xspace}
\makeatletter
\DeclareRobustCommand\onedot{\futurelet\@let@token\@onedot}
\def\@onedot{\ifx\@let@token.\else.\null\fi\xspace}
\def\eg{\emph{e.g}\onedot} 
\def\ie{\emph{i.e}\onedot} 
\def\cf{\emph{c.f}\onedot} 
 \def\vs{\emph{vs}\onedot}

\makeatother

\newcommand{\think}[1]{\textcolor{blue}{\texttt{<think>}} #1 \textcolor{blue}{\texttt{</think>}}}
\newcommand{\bbox}[1]{\textcolor{cyan}{\texttt{<box>}} #1 \textcolor{cyan}{\texttt{</box>}}}

\newcommand{\answer}[1]{\textcolor{purple}{\texttt{<answer>}} #1 \textcolor{purple}{\texttt{</answer>}}}

\title{Ground-R1: Thinking with Images via Scale Relative Policy Optimization}

\author{Meng Cao\textsuperscript{1}\footnotemark[1],~~Haoze Zhao\textsuperscript{1}\footnotemark[1],~~Can Zhang\textsuperscript{2},~~Xiaojun Chang\textsuperscript{1, 3},\\\textbf{Ian Reid}\textsuperscript{1},~~\textbf{Xiaodan Liang}\textsuperscript{1, 4}$^{\dagger}$ \\
\textsuperscript{1}Mohamed bin Zayed University of Artificial Intelligence~~\textsuperscript{2}Peking University\\~~\textsuperscript{3}University of Science and Technology of China~~\textsuperscript{4}Sun Yat-sen University\\
{\small{\textsuperscript{*}Authors contributed equally to this research.~~\textsuperscript{\dag}Corresponding author.}}\\
	\textbf{\url{https://github.com/zzzhhzzz/Ground-R1}}\\
}

\begin{document}

\twocolumn[{
	\renewcommand\twocolumn[1][]{#1}
	\maketitle
    \begin{center}
		\includegraphics[width=0.98\linewidth]{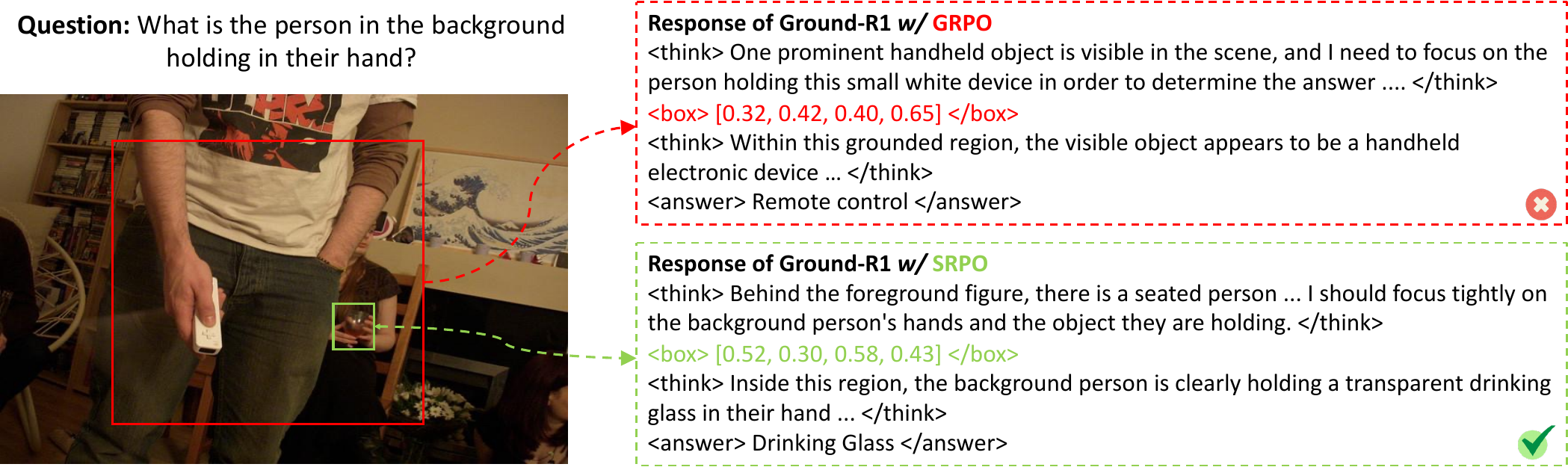} 
		\captionsetup{type=figure}
     \caption{\textbf{Comparison between standard GRPO and our proposed SRPO.} GRPO grounds on a large, salient foreground region and produces an incorrect answer by relying on spurious context. In contrast, SRPO encourages more precise evidence localization, allowing the model to attend to the small yet critical background region.}
		\label{fig:teaser}
	\end{center}
}]
\begin{abstract}
Large Vision-Language Models (LVLMs) have become powerful general-purpose assistants, yet their predictions often lack reliability and interpretability due to insufficient grounding in visual evidence. The emerging thinking-with-images paradigm seeks to address this issue by explicitly anchoring reasoning to image regions. However, we empirically find that most existing methods suffer from a systematic \emph{scale-driven bias} in optimization, where training rewards are dominated by large visual regions, suppressing learning from small but semantically critical evidence and leading to spurious grounding at inference time. To address this limitation, we propose \textbf{Ground-R1}, a de-biased thinking-with-images framework trained via a novel \textbf{S}cale \textbf{R}elative \textbf{P}olicy \textbf{O}ptimization (\textbf{SRPO}) objective that replaces standard GRPO. Specifically, our SRPO recalibrates reward learning across evidence regions of different sizes through scale-aware binning and intra-/inter-bin comparisons, enabling balanced credit assignment during training. Experimental results on general LVLM, high-resolution, and visual grounding benchmarks validate the effectiveness of Ground-R1 and show that SRPO yields consistent gains over standard GRPO in both response accuracy and evidence grounding.
\end{abstract}
\section{Introduction} \label{sec:intro}

Large Vision-Language Models (LVLMs) \cite{gpt4o,anthropic2024claude,bai2025qwen2} have emerged as versatile general-purpose assistants, capable of interpreting and executing a wide range of real-world tasks through the unified processing of visual and linguistic signals. Despite their impressive capabilities, LVLMs still suffer from limited answer reliability and poor interpretability. These issues often stem from the models' tendency to rely on spurious correlations in pre-training data distributions, rather than grounding their predictions in salient visual evidence from the input \cite{gupta2023bias,xu2024pride,zhou2024unibias}.

\begin{figure*}[t]
        \centering
         \includegraphics[width=0.96\textwidth]{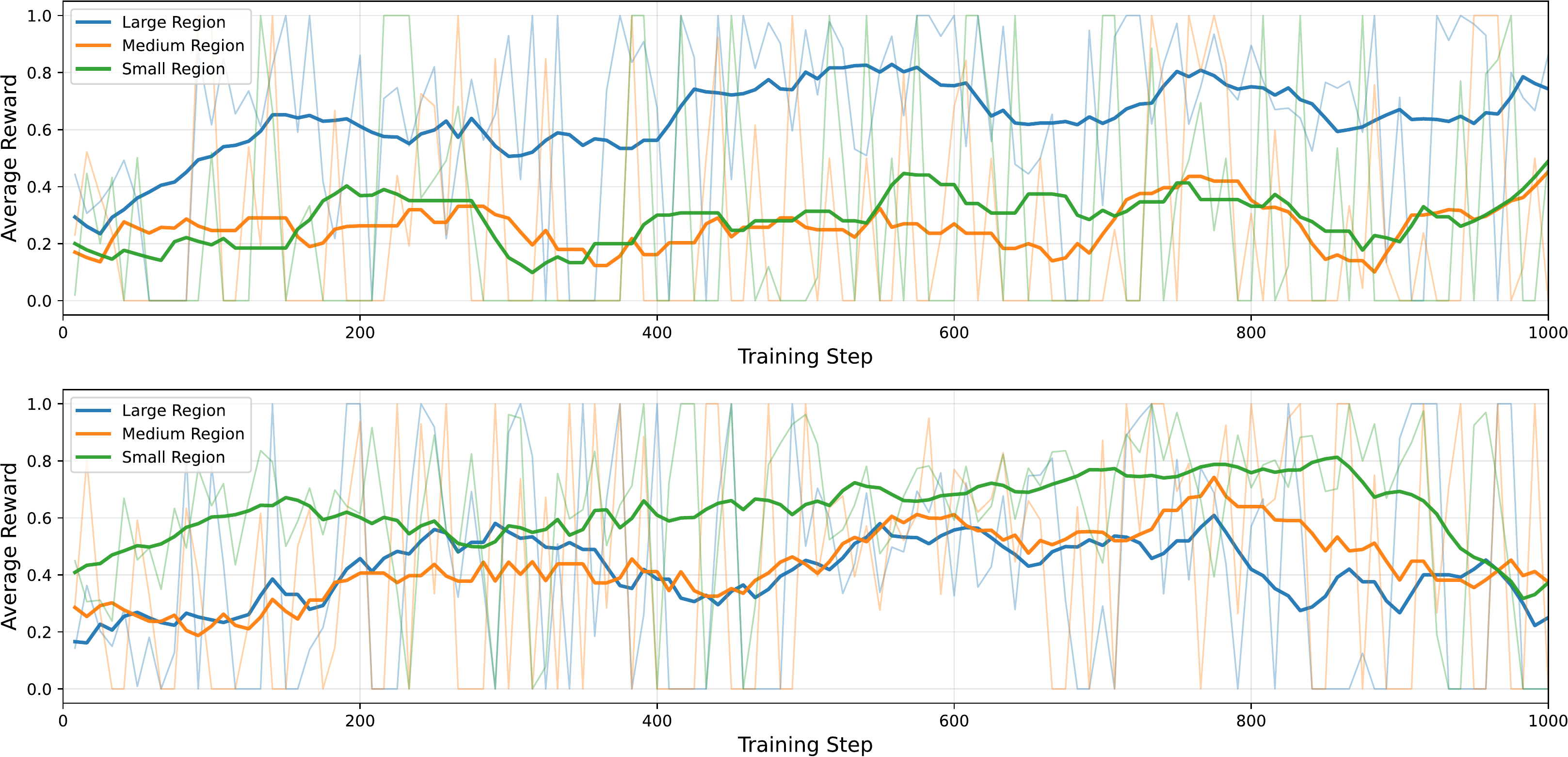}
         \vspace{-2mm}
        \caption{\textbf{Comparison of training reward trajectories} across evidence regions of different scales. \textbf{Top: vanilla GRPO}, where large regions consistently receive higher rewards. \textbf{Bottom: our proposed SRPO}, which alleviates scale-driven bias and yields more balanced reward signals across regions of different sizes. The large-, medium-, and small-scale regions are defined by relative area thresholds of less than 10\%, between 10\% and 30\%, and greater than 30\%, respectively.}
        \label{fig:regionReward}
\end{figure*}


To mitigate these limitations, recent studies advocates \emph{thinking with images} \cite{shao2024visual,su2025thinking,lai2025mini,zheng2025deepeyes}, \ie, a paradigm that requires models to explicitly anchor reasoning processes to semantically grounded evidence regions. Despite this progress, most existing approaches build upon vanilla GRPO \cite{guo2025deepseek}, which relies on outcome-level rewards without step-wise supervision, making accurate evidence grounding difficult to ensure. Empirically, we further identify a systematic \emph{scale-driven bias}, where the training rewards are dominated by large visual regions. To substantiate this observation, we conduct diagnostic experiments to analyze reward distributions across evidence regions of different scales during training. Specifically, we track reward trajectories for large-, medium-, and small-scale regions, defined by relative area thresholds of less than 10\%, between 10\% and 30\%, and greater than 30\%, respectively. As shown in Figure \ref{fig:regionReward}, the large evidence regions consistently receive higher rewards than medium and small ones, indicating that they dominate the training signal. This leads to consistently negative advantages for small regions with lower mean rewards, causing their gradients to be suppressed or even clipped during optimization. Consequently, at inference time, models tend to favor visually prominent objects over small yet semantically critical evidence (\cf Figure \ref{fig:teaser}).

To address this issue, we propose \textbf{Ground-R1}, a de-biased thinking-with-images framework equipped with \textbf{S}cale \textbf{R}elative \textbf{P}olicy \textbf{O}ptimization (\textbf{SRPO}), a novel training paradigm that recalibrates reward signals across evidence regions of different scales. Specifically, SRPO first discretizes evidence regions into scale-aware bins based on their relative areas. It then decomposes reward learning into complementary \emph{intra-bin} and \emph{inter-bin} components, capturing scale-calibrated relative comparisons within each scale while encouraging discriminative comparisons across scales. Furthermore, SRPO adopts a scale-aware advantage computation strategy, where intra-bin advantages are computed using per-bin statistics and inter-bin advantages are estimated globally. This scale-relative design effectively mitigates the dominance of large regions in training, leading to more balanced reward propagation and more faithful visual evidence grounding.

We evaluate Ground-R1 across a comprehensive set of benchmarks covering general LVLM tasks, high-resolution scenarios, and visual grounding evaluations. Our method consistently delivers strong performance over supervised fine-tuning (SFT) counterparts, grounded visual reasoning approaches \cite{liu2024chain,shao2024visual,qi2024cogcom,liu2024chain}, R1-series works \cite{yang2025r1,peng2025lmm,huang2025vision} and concurrent thinking-with-image methods \cite{zheng2025deepeyes,lai2025mini}. For example, built upon Qwen2.5-VL \cite{bai2025qwen2}, Ground-R1 achieves a 11.9\% absoluate improvement on V$^{*}$ \cite{wu2024v}. In addition, compared to standard GRPO, our proposed SRPO consistently delivers further performance gains while enabling more precise and faithful grounding of evidence regions. Specifically, Ground-R1 trained via SRPO achieves an 1.8\% absolute improvement over the GRPO variant on the HR-8K benchmark. Moreover, SRPO demonstrates improved evidence region grounding over the course of training, resulting in higher grounding IoU values than GRPO (\cf Figure \ref{fig:visCurve}).

In summary, our contributions are in three-folds:
\begin{itemize}[topsep=0pt, partopsep=0pt, leftmargin=20pt, parsep=0pt, itemsep=3pt]
    \item We introduce Ground-R1, a thinking-with-images paradigm that decomposes reasoning into grounding and answering stages to explicitly anchor predictions in visual evidence.

    \item We identify and systematically diagnose a scale-driven bias in GRPO-based grounded reasoning, and propose scale relative policy optimization to recalibrate rewards across evidence regions of varying sizes.

    \item Extensive experiments across general LVLM, high-resolution, and visual grounding benchmarks demonstrate that Ground-R1 consistently outperforms prior methods, with SRPO yielding consistent gains over GRPO in both answering accuracy and evidence grounding.
\end{itemize}

\section{Related Work}\label{relatedwork}

\noindent \textbf{Visual Reasoning in LVLMs.} Recent advances in LVLMs' visual reasoning capabilities have driven sustained research efforts toward implementing cognitive processes through explicit problem-solving trajectories \cite{wang2025multimodal,huang2023towards,xu2025towards}. The early attempts \cite{chen2023see,yao2023tree,besta2024graph,wei2022chain} employ carefully designed prompts to guide the rationale generation. + Another line of work enhances LVLM reasoning capabilities through supervised fine-tuning on rationale annotations \cite{zhang2023multimodal,wang2024t,dong2024insight,xu2024llava,thawakar2025llamav,yao2024mulberry}. LLaVA-CoT \cite{xu2024llava} explicitly defines the structured reasoning stages to enhance the process interpretability. Mulberry \cite{yao2024mulberry} augments Monte Carlo Tree Search by integrating collective knowledge from multiple LVLMs, where negative exploration paths are leveraged to synthesize self-reflective data for iterative self-correction. Recent breakthroughs from DeepSeek-R1 \cite{guo2025deepseek}  have catalyzed a wave of R1-inspired methodologies in vision-language research \cite{yang2025r1,peng2025lmm,tan2025reason,liu2025visual,meng2025mm,yu2025perception,huang2025vision}. Vision-R1 \cite{huang2025vision} introduces a progressive thinking suppression training strategy to address overthinking issues. LMM-R1 \cite{peng2025lmm} focuses on unleashing general reasoning capabilities in 3B-parameter models, demonstrating progress in visual geometry and agent domains. Visual-RFT \cite{liu2025visual} further extends this paradigm to broader vision tasks, including fine-grained classification and object detection. Despite recent progress, most LVLMs still adopt a text-centric reasoning paradigm, whereas human cognition relies on interleaved visual–text reasoning grounded in visual evidence, motivating the need for more grounded reasoning frameworks.

\noindent \textbf{Thinking with Images.} Recent advances in multimodal reasoning have moved beyond text-centric paradigms toward thinking with images \cite{shao2024visual,su2025thinking,lai2025mini,zheng2025deepeyes}, where visual information is actively involved as intermediate representations during reasoning rather than serving as static context. Existing methods can be broadly categorized into three types according to how visual information is manipulated: 1) \emph{Tool-driven reasoning} \cite{wu2024mind,liu2023llava,qi2024cogcom,shao2024visual,hong2025deepeyesv2,zheng2025deepeyes,lai2025mini}, where models act as high-level planners that orchestrate predefined visual tools, \eg, visual, numerical, and information retrieval tools, to optimize grounded reasoning policies; 2) \emph{Programmatic reasoning} \cite{gupta2023visual,suris2023vipergpt,liu2025visual}, which treats visual reasoning as a code generation process, allowing models to compose customized and verifiable visual operations, particularly effective for structured tasks such as geometry \cite{mallis2025cad} and mathematical reasoning \cite{wang2025mathcoder}; 3) \emph{Intrinsic reasoning} \cite{team2024chameleon,sun2024generative,chen2025blip3,xu2025visual}, where models internally synthesize visual representations within a closed reasoning loop, unifies visual generation and reasoning by making visual imagination a native component of the reasoning process. Our Ground-R1 follows the tool-driven reasoning paradigm, avoiding the computational overhead of programmatic execution and the uncertainty of intrinsic visual imagination \cite{hu2024visualProg}. Moreover, we identify a systematic bias in reward learning and propose SRPO as an alternative to GRPO that recalibrates reward signals in a scale-aware manner.
\section{Ground-R1} \label{sec:3}

\begin{figure*}[t]
	\centering
        \includegraphics[width=\textwidth]{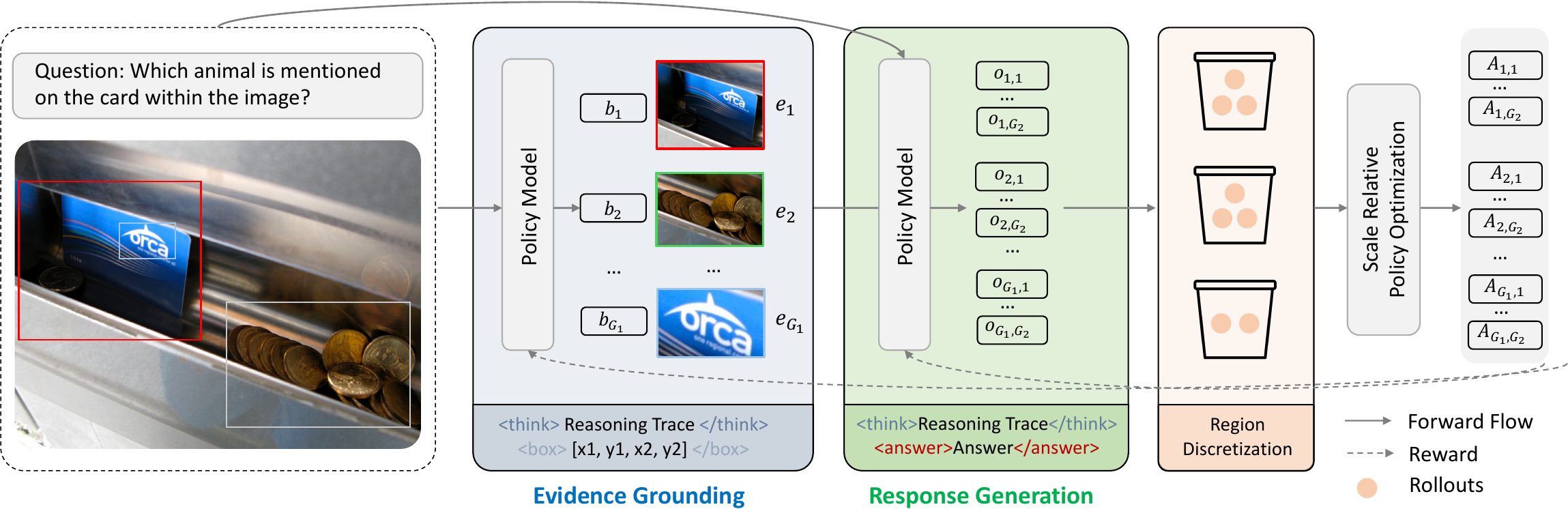}
	\caption{\textbf{Schematic illustrations of Ground-R1. The grounding phase} analyzes input instructions and generates evidence region rollouts. $\boldsymbol{b}_i \in \mathbb{R}^{4}$ denotes the axis-aligned bounding box coordinates and $\boldsymbol{e}_i$ is the corresponding cropped evidence region, $i \in [1, G_1]$. \textbf{The answering phase} takes the input image, question, and the generated evidence regions as input and delivers final answers. $\boldsymbol{o}_{i,j}$ denote the $j$-th rollout answers based on the $i$-th evidence region $\boldsymbol{e}_{i}$. $j \in [1, G_2]$. \textbf{Region discretization} partitions rollouts into $K$ equal-sized bins based on their relative areas. $A_{i,j}$ is the computed advantages (\cf Eq. \eqref{eq:5}) and the proposed SGPO is optimized via Eq. \eqref{eq:6}.}
	\label{fig:pipeline}
\end{figure*}

A schematic illustration of Ground-R1 is shown in Figure \ref{fig:pipeline}. The overall pipeline is described in Section \ref{sec:3.1}, while the proposed scale-relative policy optimization is detailed in Section \ref{sec:3.2}.

\subsection{Overall Pipeline} \label{sec:3.1}

\noindent \textbf{Grounding Rollout.} Given the input question $\boldsymbol{q}$ and image $\boldsymbol{v}$, the grounding rollout process synthesizes critical evidence regions through iterative reasoning. We formally structure cognitive processes through dedicated markup tags: analytical reasoning flows are encapsulated within \think{and} elements, while visual grounding outputs are delimited by \bbox{and} tags. 

Following GRPO \cite{shao2024deepseekmath}, a set of grounding rollouts is generated by an existing policy $\pi_{\theta_{\text{old}}}$ as follows.
\begin{equation}
\boldsymbol{b}=\left\{\boldsymbol{b}_i\right\}_{i=1}^{G_1} \sim \pi_{\theta_{\text {old }}}(\cdot \mid \boldsymbol{q}, \boldsymbol{v}),
\end{equation}
\noindent where $\boldsymbol{b}_i \in \mathbb{R}^{4}$ denotes the axis-aligned bounding box represented by its top-left and bottom-right corner coordinates. $G_1$ is the number of sampled grounding rollouts.

\noindent \textbf{Answer Rollout.} Leveraging the evidence region coordinates generated from the grounding rollout process, we extract corresponding image regions through spatial cropping. During the answer rollout phase, the policy model integrates multiple inputs including the original image, textual query, and evidence-aligned image regions to synthesize final responses. Consider a generated bounding box $\boldsymbol{b}_i$ as an illustrative case, the corresponding evidence region $\boldsymbol{e}_i$ is derived through zoom-in and cropping operations. The answer rollout process is formally expressed as follows:
\begin{equation}
\boldsymbol{o}_i=\left\{\boldsymbol{o}_{i,j}\right\}_{j=1}^{G_2} \sim \pi_{\theta_{\text {old }}}(\cdot \mid \boldsymbol{q}, \boldsymbol{v}, \boldsymbol{e}_{i}), \label{eq:2}
\end{equation}
\noindent where $\boldsymbol{o}_{i,j}$ denote the $j$-th rollout answers based on the $i$-th evidence region $\boldsymbol{e}_{i}$. $i \in [1, G_1]$, $j \in [1, G_2]$. $G_2$ denotes the number of answer rollouts generated per evidence region, yielding a total of $G_1 \cdot G_2$ reasoning trajectories.




\subsection{Scale Relative Policy Optimization} \label{sec:3.2}
Our Ground-R1 is trained via the proposed SRPO, which extends the conventional GRPO \cite{shao2024deepseekmath} by conditioning advantage normalization on the region scale. 

\noindent \textbf{Evidence Region Discretization.} We first perform the \emph{area-based discretization} on the evidence regions $\boldsymbol{e}_{i}$ generated in the grounding rollout process, partitioning them into $K$ equal-sized bins based on their relative areas, as follows:
\begin{equation}
\boldsymbol{s}(i) = \min\left(\left\lfloor K \boldsymbol{a}_i \right\rfloor + 1, K\right) \in \{1, \dots, K\}. \label{eq:3}
\end{equation}
\noindent where $\boldsymbol{s}(i)$ denotes the bucket index of the $i$-th evidence region and $\boldsymbol{a}_i$ is the relative area of $\boldsymbol{e}_i$.

\noindent \textbf{Reward Scoring.} Since the rewards vary significantly across different area bins, we apply binary rewards including both \emph{intra-} and \emph{inter-bin rewards}.
\begin{itemize}[topsep=0pt, partopsep=0pt, leftmargin=10pt, parsep=0pt, itemsep=3pt]
   \item \emph{Intra-bin reward}: It is designed to capture the relative magnitudes within each bin. Specifically, it incorporates two complementary components including a \emph{format reward} which ensure that the responses conform to the desired format and an \emph{answer reward} which evaluates the correctness of the final prediction. Formally, the intra-bin reward is denoted as $r^{\text{intra}}_{i,j}$, $i \in [1, G_1]$, $j \in [1, G_2]$. For each bin $\boldsymbol{S}_k = \{(i, j) \mid \boldsymbol{s}(i) = k\}$, $k \in [1, K]$, we compute the mean and standard deviation of the intra-bin rewards, denoted as $\mu_k$ and $\sigma_k$, respectively.

   \item  \emph{Inter-bin reward}: To encourage discriminative comparisons across different bins, we introduce an additional inter-bin reward that favors rollouts associated with the bin exhibiting the highest mean reward. Specifically, the inter-bin reward $r^{\text{inter}}_{i,j}$ is computed as follows.
    \begin{equation}
    r^{\text{inter}}_{i,j}= \begin{cases}1, & \text {if } i=\underset{k \in\{1, \ldots, K\}}{\arg \max } \mu_k \\ 0, & \text {otherwise }\end{cases} \label{eq:4}
    \end{equation}      
\end{itemize}

\noindent \textbf{Scale-aware Advantage Estimation.} For intra-bin rewards, advantages are estimated using \emph{per-bin} statistics instead of global normalization to ensure scale-consistent within each bin. By contrast, inter-bin rewards capture cross-bin discrimination and thus use standard \emph{global} advantage estimation.
\begin{equation}
A_{i,j} = \frac{r^{\text{intra}}_{i,j} - \mu_{s(i)}}{\sigma_{s(i)}} +  \frac{r^{\text{inter}}_{i,j} - \text{mean}(r^{\text{inter}}_{i,j})}{\text{std}(r^{\text{inter}}_{i,j})}, \label{eq:5}
\end{equation}
\noindent  where $\mu_{s(i)}$ and $\sigma_{s(i)}$ denote the mean and standard deviation of the intra-bin rewards for the bin to which the $i$-th evidence region belongs.

\noindent \textbf{Optimization Objective.} Our proposed SRPO is optimized by maximizing the following objective:
\begin{equation}
\small
\begin{gathered}
\mathcal{J}_{}(\theta)=\mathbb{E}_{\boldsymbol{q}, \boldsymbol{v},\left\{\boldsymbol{o}_{i,j}\right\}_{i=1,j=1}^{G_1,G_2} \sim \pi_{\theta_{\text {old}}}}\bigg[ \frac{1}{G_1G_2} \sum_{i=1}^{G_1}\sum_{j=1}^{G_2}\\
\min \Big(\rho_{i,j}(\theta) A_{i,j}, \;\operatorname{clip}\big(\rho_{i,j}(\theta), 1-\varepsilon, 1+\varepsilon\big) A_{i,j}\Big)\\
- \beta \, \mathbb{D}_{\mathrm{KL}}(\pi_\theta \| \pi_{\mathrm{ref}})\bigg],
\end{gathered}
\label{eq:6}
\end{equation} 
\noindent where $\varepsilon$ is clipping-related hyper-parameters for stabilizing training and the importance sampling ratio $\rho_{i, j}(\theta)$ is defined as follow:
\begin{equation}
\rho_{i, j}(\theta)=\frac{\pi_\theta\left(\boldsymbol{o}_{i, j} \mid \boldsymbol{q}, \boldsymbol{o}_{i,<j}\right)}{\pi_{\theta_{\text {old }}}\left(\boldsymbol{o}_{i, j} \mid \boldsymbol{q}, \boldsymbol{o}_{i,<j}\right)}.
\end{equation}

\section{Experiments} \label{sec:4}

\subsection{Experimental Settings}  \label{sec:4.1}
\begin{table*}[t]
\belowrulesep=0pt
\aboverulesep=0pt
    \centering
    \renewcommand\arraystretch{1.1}
    \setlength{\tabcolsep}{3mm}
    \caption{\textbf{Evaluation results on LVLM benchmarks} including general and high-resolution scenarios. HR-4K/8K denotes HR-Bench \cite{wang2025divide} with the high resolution of 4K/8K. MME-RWL denotes the MME-RealWorld lite version \cite{zhang2024mme}. $^{\dagger}$ denotes results reproduced by us under the same experimental setting.}
    \resizebox{0.98\textwidth}{!}{
    \begin{tabular}{c|cccccccccc}
    \toprule
    \rowcolor{gray!10} 
    \textbf{Method} & \textbf{MME} & \textbf{MM-Vet} & \textbf{SEED} & \textbf{MME-RWL} & \textbf{RWQA} & \textbf{POPE} & \textbf{V$^{*}$} & \textbf{HR-4K}  & \textbf{HR-8K} \\
    \hline
    GPT-4V \cite{gpt4v} & 1926.6 & 67.5 & 53.8 & -- & 61.4 & -- & -- & -- & -- \\
    GPT-4o \cite{gpt4o} & -- & 69.1 & 72.0 & -- & 75.4 & 86.9 & 65.2 & 62.0  & 58.3\\
    InternVL2-8B \cite{chen2024far} & 2210.3 & 54.2 & -- & -- & 64.4 & 86.9 & -- & -- & -- \\
    InternVL2.5-8B \cite{chen2024expanding} & 2344.1 & 62.8 & -- & -- & 70.1 & 90.6 & -- & -- & -- \\
    Qwen2-VL-7B \cite{wang2024qwen2} & 2326.8 & 62.0 & 75.1 & -- & 70.1 & 88.1 & -- & -- & -- \\
    CogCoM \cite{qi2024cogcom} & -- & 46.1 & -- & -- & -- & -- & -- & -- & -- \\
    Chain of Spot  \cite{liu2024chain}& 1501.1 & 30.8 & 59.7 & -- & -- & 86.4 & -- & -- & -- \\
    FaST \cite{sun2024visual} & 1517.4 & 31.0 & 60.1 & -- & -- & 86.4 & -- & -- & -- \\
    Vision-R1-7B \cite{huang2025vision} & 2306.2 & 58.3 & 73.0 & 41.0 & 50.2 & 88.7 & -- & -- & -- \\
    LMM-R1 \cite{peng2025lmm} & 2210.6 & 64.5 & 74.4 & 35.4 & 52.2 & 86.5 & -- & -- & -- \\
    R1-Onevision \cite{yang2025r1} & 2192.2 & 67.5 & 66.5 & 35.1 & 46.5 & 84.9 & -- & -- & -- \\
    LLaVA-OneVision \cite{li2024llava} & -- & -- & -- & -- & -- & -- & 75.4 &  63.0 & 59.8  \\
    DeepEyes$^{\dagger}$ \cite{zheng2025deepeyes} & -- & -- & -- & -- & -- & 87.7 &  83.3 & 73.2 & 69.5  \\
    Mini-o3$^{\dagger}$ \cite{lai2025mini} & -- & -- & -- & -- & -- & -- & 86.9 & 74.9 & 70.6 \\
    Pixel Reasoner \cite{wang2025pixel} & -- & -- & -- & -- & -- & -- & 86.3 & 74.0 & 66.9 \\
    \hline 
    Qwen2.5-VL-7B \cite{bai2025qwen2} & 2327.0 & 67.1 & 76.5 & 33.8 & 65.9 & 83.6 & 75.5 & 68.2 & 62.7\\
    Ground-R1 (Ours) & \textbf{2410.4} & \textbf{69.8} & \textbf{77.8} & \textbf{50.9} & \textbf{68.8} & \textbf{89.8} & \textbf{87.4} & \textbf{75.0} & \textbf{71.1}  \\
    \(\Delta\) (\vs Qwen2.5-VL-7B) & +83.4 & +2.7 & +1.3 & +17.1 & +2.9 & +6.2 & +11.9 & +6.8 & +8.4 \\
    \bottomrule
    \end{tabular}
    }
    \label{tab:resGeneral}
\end{table*}
\noindent \textbf{Training Dataset.} We adopt the dataset introduced by DeepEyes \cite{zheng2025deepeyes} including fine-grained visual search \cite{wu2024v}, arXivQA \cite{li2024multimodal}, ThinkLite-VL \cite{wang2025sota} for RL training. Note that we do not employ curated SFT dataset for cold-start training, as the distribution of SFT data can substantially alter the grounding behavior of LVLMs \cite{lai2025mini}, which would hinder our investigation of the base model's inherent behavior under the thinking-with-images paradigm.

\noindent \textbf{Implementation Details.} We conducted experiments using Qwen-2.5-VL-7B-Instruct \cite{bai2025qwen2}. The training was performed for 1,000 steps with a batch size of 8 and a learning rate of {$1\times e^{-6}$}. The rollout numbers for the grounding ($G_1$) and answering ($G_2$) stages were set to 4 and 2, respectively. The number of equal-sized bins $K$ was set to 3. We used a default sampling temperature of 1 and a maximum response length of 512 tokens. All experiments were run on 8$\times$H100 GPUs, with a total training time of approximately 12 hours.

\subsection{Evaluation on LVLM Benchmarks.}  \label{sec:4.2}
\noindent \textbf{Results on General Benchmarks.} Table \ref{tab:resGeneral} reports the quantitative comparisons on six widely used benchmarks, including MME \cite{fu2306mme}, MM-Vet \cite{yu2023mm}, SEED-Bench \cite{li2023seed}, MME-RealWorld-Lite \cite{zhang2024mme}, RealworldQA \cite{xai2024grok15v}, and POPE \cite{li2023evaluating}. These benchmarks jointly evaluate general multi-modal reasoning and hallucination resistance. We include strong proprietary models (\eg, GPT-4V and GPT-4o), state-of-the-art open-source LVLMs (\eg, InternVL2.5, Qwen2-VL), and recent thinking-with-images (\eg, DeepEyes, Mini-o3) and R1-style approaches (\eg, Vision-R1, LMM-R1) for comprehensive comparison.

As shown in Table \ref{tab:resGeneral}, Ground-R1 consistently outperforms its base model Qwen2.5-VL-7B across all evaluated benchmarks, \eg, achieving notable absolute improvements of +83.4 on MME and +17.1\% on MME-RWL. These results indicate that Ground-R1 substantially enhances both general multimodal reasoning and grounding-sensitive capabilities without sacrificing robustness.

\noindent \textbf{Results on High-Resolution Benchmarks.} We provide the comparison results on high-resolution benchmarks, including V$^{}$ \cite{wu2024v} and HR-Bench \cite{wang2025divide}, which pose more challenging evaluations due to high-resolution images (4K/8K) and the presence of small-scale objects. As shown, Ground-R1 consistently outperforms Qwen2.5-VL-7B as well as prior methods across all high-resolution settings, achieving notable absolute improvements of +11.9\% on V*, +6.8\% on HR-4K, and +8.4\% on HR-8K. These gains indicate that Ground-R1 is particularly effective over high-resolution scenarios where critical evidence often occupies a small spatial extent.

\subsection{Visual Grounding}  \label{sec:4.3}

To ensure the broad coverage of multi-modal competencies, we also evaluate the visual grounding capabilities of Ground-R1 on RefCOCO \cite{kazemzadeh2014referitgame}, RefCOCO+ \cite{kazemzadeh2014referitgame} and RefCOCOg \cite{mao2016generation}. We use the following prompts as follows, \ie, ``\emph{Locate} \emph{<ref>} \emph{in this image and output the bbox coordinates in JSON format.}", where \emph{<ref>} refers to the specific expression. 

As shown in Table \ref{tab:resVisGround}, Ground-R1 demonstrates superior visual grounding capabilities, achieving 93.1\% accuracy on RefCOCO val and surpassing most LVLMs, while narrowing the performance gap between generalist and specialist models (\eg, Grounding DINO \cite{liu2024grounding}). Although Ground-R1 slightly underperforms CogCoM \cite{qi2024cogcom} on specific benchmarks (\eg, test-A of RefCOCO), this discrepancy is explainable: CogCoM leverages a specialized grounded VQA dataset \cite{wang2024cogvlm} for additional training to enhance visual grounding, whereas our Ground-R1 maintains dual competence in both visual reasoning and grounding without requiring dedicated grounding-specific training.

\begin{table*}[t]
\belowrulesep=0pt
\aboverulesep=0pt
    \centering
    \setlength{\tabcolsep}{5mm}
     \renewcommand\arraystretch{1.1}
    \caption{\textbf{Evaluation results on visual grounding benchmarks} including RefCOCO, RefCOCO+ and RefCOCOg.}
    \vspace{-2mm}
    \resizebox{0.95\textwidth}{!}{
    \begin{tabular}{c|ccc|ccc|cc}
    \toprule
    \rowcolor{gray!10}
    &  \multicolumn{3}{c|}{\textbf{RefCOCO}} & \multicolumn{3}{c|}{\textbf{RefCOCO+}} & \multicolumn{2}{c}{\textbf{RefCOCOg}} \\ \rowcolor{gray!10}
    \multirow{-2}{*}{\textbf{Method}}  & val  & test-A & test-B & val & test-A & test-B & val & test\\
    \hline
    Gemini 1.5 Pro \cite{reid2024gemini} & 73.2 & 72.9 & 74.6 & 62.5 & 63.9 & 65.0 & 75.2 & 76.2\\
    Grounding DINO \cite{liu2024grounding}  & 90.6 & 93.2 &  88.2 & 88.2 & 89.0 &  75.9 &  86.1 & 87.0  \\
    Shikra-7B  \cite{chen2023shikra} & 87.0 & 90.6 & 80.2 & 81.6 & 87.4 & 72.1 & 82.3 & 82.2 \\
   Shikra-13B  \cite{chen2023shikra} & 87.8 & 91.1 & 81.8 & 82.9 & 87.8 & 74.4 & 82.6 & 83.2 \\
   Qwen2-VL  \cite{bai2023qwen}   & 89.4 & 92.3 & 85.3 & 83.1 & 88.3 & 77.2 & 85.6 & 85.5\\
   CogVLM \cite{wang2024cogvlm}  & 92.5 & 94.0 & 88.7 & 87.5 & 91.8 & 81.4 & 89.5 & 90.1 \\
   CogCoM \cite{qi2024cogcom} & 92.3 & 94.6 & 89.2 & 88.2 & 92.8 & 82.1 & 89.3 & 90.5 \\
   Qwen2.5-VL-7B \cite{bai2025qwen2} & 90.0 & 92.5 & 85.4 & 84.2 & 89.1 & 76.9 & 87.2 & 87.2 \\
   InternVL2.5-8B \cite{chen2024expanding} & 90.3 & 94.5 & 85.9 & 85.2 & 91.5 & 78.8 & 86.7 & 87.6 \\
   Vision-R1-7B \cite{huang2025vision} & 58.3 & 64.2 & 48.5 & 50.9 & 61.1 & 38.5 & 51.9 & 52.6  \\
    LMM-R1 \cite{peng2025lmm} & 87.3 & 91.4 & 83.6 & 79.2 & 87.1 & 72.9 & 84.1 & 84.1 \\
    R1-Onevision \cite{yang2025r1} & 45.0 & 51.6 & 35.3 & 47.2 & 57.0 & 37.1 & 53.0 & 48.2\\
    DeepEyes \cite{zheng2025deepeyes} & 89.8 & -- & -- & 83.6 & -- &  -- & 86.7 &  --\\
    \hline 
    Qwen2.5-VL-7B \cite{bai2025qwen2} & 90.0 & 92.5 & 85.4 & 84.2 & 89.1 & 76.9 & 87.2 & 87.2 \\
    Ground R1 (Ours) & \textbf{93.1} & \textbf{94.0} & \textbf{88.2} & \textbf{86.9} & \textbf{91.2} & \textbf{78.8} & \textbf{89.8} & \textbf{90.2}    \\
    \(\Delta\) (\vs Qwen2.5-VL-7B) & +3.1 & +1.5 & +2.8 & +2.7 & +2.1 & +1.9 & +2.6 & +3.0  \\
    \bottomrule
    \end{tabular}}
    \label{tab:resVisGround}
\end{table*}

\begin{table*}[t]
\belowrulesep=0pt
\aboverulesep=0pt
    \centering
        \setlength{\tabcolsep}{3mm}
    \renewcommand\arraystretch{1.1}
    \caption{\textbf{Ablation studies of Ground-R1.} Refer to Section \ref{sec:4.4} for the configuration of each model variants.}
    \vspace{-2mm}
    \resizebox{0.96\textwidth}{!}{
    \begin{tabular}{c|cccccccccccc}
    \toprule
    \rowcolor{gray!10} 
    \textbf{Method} & \textbf{MME} & \textbf{MM-Vet} & \textbf{SEED} & \textbf{MME-RWL} & \textbf{RWQA} & \textbf{POPE} & \textbf{V$^{*}$} & \textbf{HR-4K}  & \textbf{HR-8K} \\
    \hline
    Ground-R1 & \textbf{2410.4} & \textbf{69.8} & \textbf{77.8} & \textbf{50.9} & \textbf{68.8} & \textbf{89.8} & \textbf{87.4} & \textbf{75.0} & \textbf{71.1}  \\ 
    Vanilla-SFT    & 2337.0 & 68.0 & 76.7 & 34.2 & 67.1 & 87.6 & 73.3 & 66.6 & 60.0 \\
    Vanilla-R1  & 2338.8 & 53.2 & 75.4 & 35.1 & 66.5  & 88.4 & 80.1 & 71.3 & 67.9 \\
    Ground-R1-GRPO & 2395.1 & 68.8 & 77.6 & 49.4 & 67.7 & 89.7 & 85.3 & 73.8 & 69.3\\
    Ground-R1-Intra & 2358.7 & 59.2 & 76.7 & 43.6 & 63.9 & 88.8 & 81.7 & 71.8 & 68.5 \\
    Ground-R1-Inter & 2390.5 & 67.9 & 76.9 & 44.1 & 65.5 & 89.3 & 81.2 & 72.6 & 68.8\\
    \bottomrule
    \end{tabular}}
    \label{tab:ablate}
\end{table*}

\subsection{Ablation Studies}  \label{sec:4.4}

To justify the design of our approach, we compare Ground-R1 against the following model variants:
\begin{itemize}[topsep=0pt, partopsep=0pt, leftmargin=10pt, parsep=0pt, itemsep=3pt]
    
    \item \textbf{Vanilla-SFT}: This variant replaces the RL-based training in Ground-R1 with SFT.
    
    \item \textbf{Vanilla-R1}: This variant omits the grounding phase of Ground-R1 and leverages RL training.
    
    \item \textbf{Ground-R1-GRPO}: This variant replaces the proposed SRPO with the standard GRPO.

    \item \textbf{Ground-R1-Intra/Inter}: These two variants use only the intra-bin or inter-bin rewards in SRPO, respectively (\cf Section \ref{sec:3.2}).
\end{itemize}

The comparison results are demonstrated in Table \ref{tab:ablate}. Building upon these model variants, we seek to address four critical inquiries:

\noindent \textbf{\emph{Q1:}} \emph{Does RL-based methodology outperform SFT-based approaches in visual reasoning?}

By comparing Ground-R1 with Vanilla-SFT, we observe that RL-based methods consistently outperform SFT-based counterparts across all the benchmarks. For example, Ground-R1 achieves an absolute improvement of 14.1\% over Vanilla-SFT on V$^{*}$ Bench. These results indicate that RL-based optimization more effectively aligns model behavior with task objectives by providing direct feedback on reasoning outcomes.

\noindent \textbf{\emph{Q2:}} \emph{Does the ``grounding-then-answering" paradigm outperform direct answer generation?}

To answer this question, we compare Ground-R1 with Vanilla-R1, where the latter removes the explicit grounding stage and directly generates answers. As shown in Table \ref{tab:ablate}, Ground-R1 consistently outperforms Vanilla-R1 across all the evaluated benchmarks. 
These results indicate that explicitly separating grounding from answering enables the model to localize task-relevant visual evidence more effectively, leading to more accurate reasoning than direct answer generation.

\noindent \textbf{\emph{Q3:}} \emph{To what extent does the proposed SRPO outperform standard GRPO?}

To quantify the effectiveness of SRPO, we compare Ground-R1 with Ground-R1-GRPO, which shares the same architecture but replaces SRPO with standard GRPO. As shown in Table \ref{tab:ablate}, Ground-R1 consistently outperforms Ground-R1-GRPO across all evaluated benchmarks. Notably, SRPO brings substantial gains on high-resolution benchmarks, achieving an improvement of +2.1\% on V$^{*}$, +1.2\% on HR-4K, and +1.8\% on HR-8K. These results demonstrate that SRPO provides additional benefits over standard GRPO by more effectively recalibrating reward signals across evidence regions of different scales.

\begin{figure*}[t]
	\centering
        \includegraphics[width=0.96\textwidth]{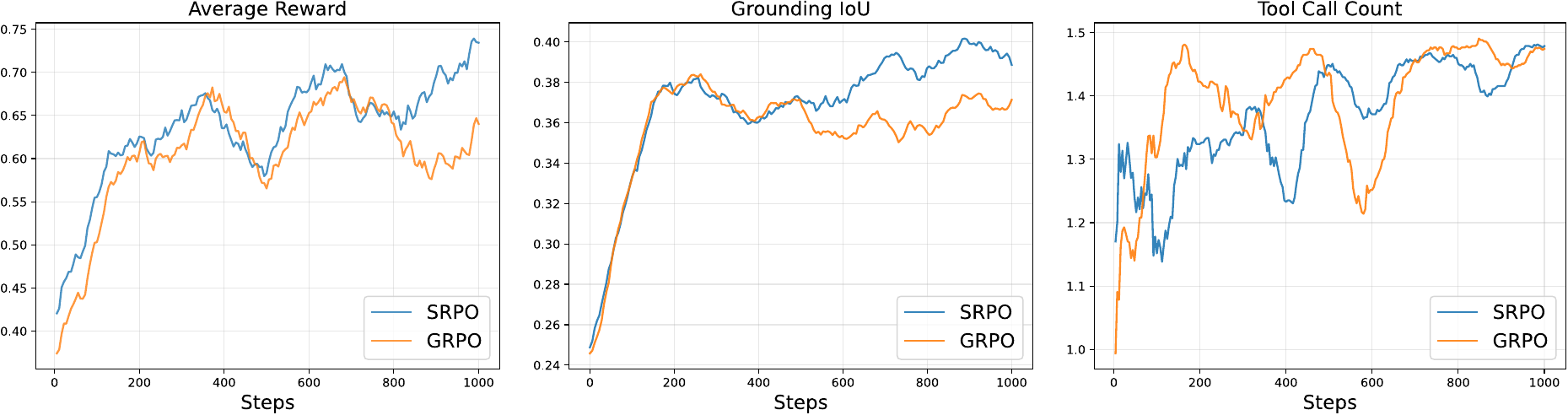}
       \vspace{-1mm}
	\caption{\textbf{Training dynamics of Ground-R1} including normalized average reward, grounding IoU, and tool call count under standard GRPO and our proposed SRPO.}
	\label{fig:visCurve}
\end{figure*}

\begin{figure*}[ht]
	\centering
        \includegraphics[width=0.92\textwidth]{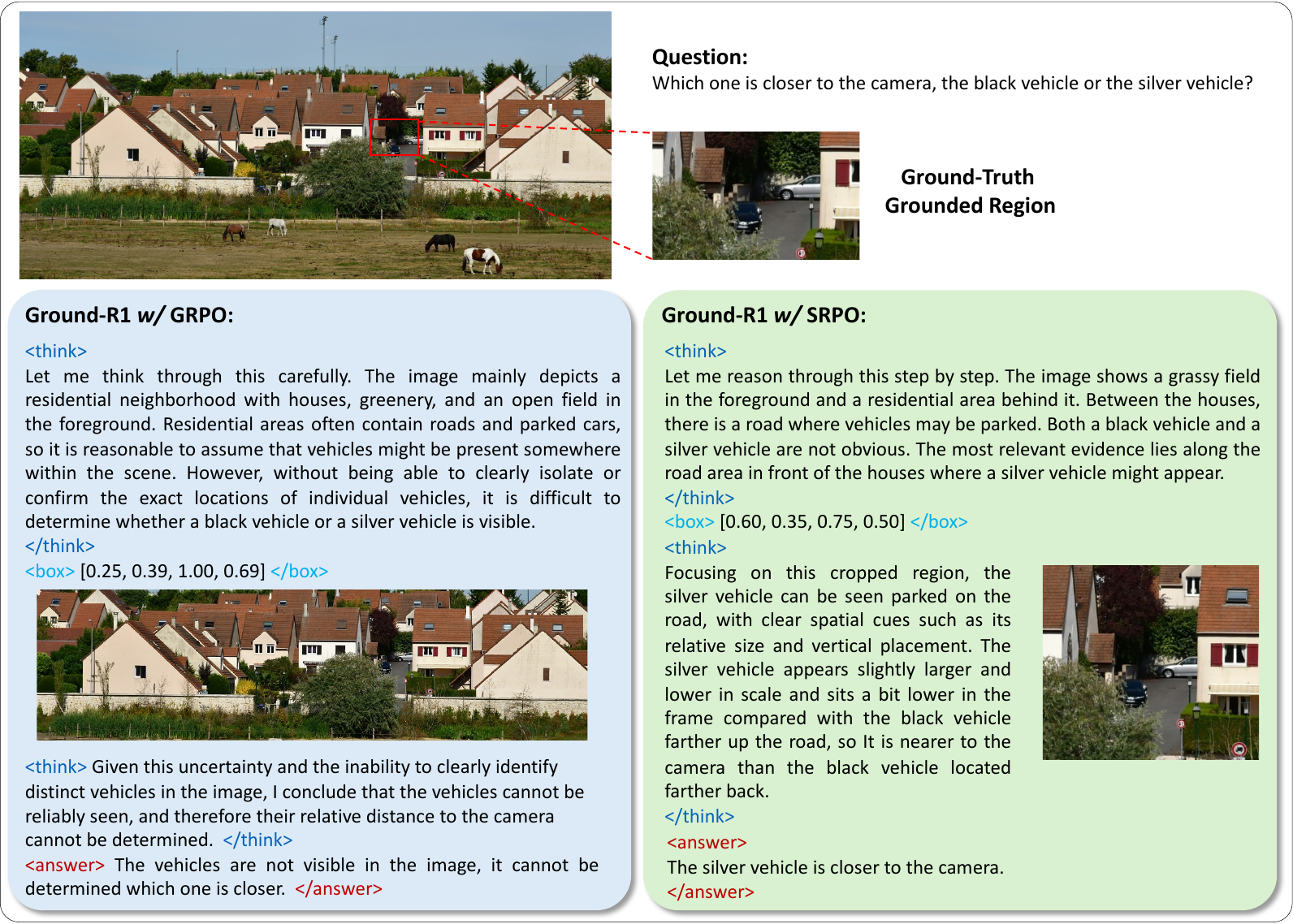}
       \vspace{-1mm}
	\caption{\textbf{Comparisons of reasoning trajectories} of Ground-R1 under standard GRPO and our proposed SRPO.}
	\label{fig:inferCompare}
\end{figure*}


\noindent \textbf{\emph{Q4:}} \emph{How do the intra- and inter-bin reward components of SRPO contribute to the performance?}

To assess the contribution of each reward component in SRPO, we compare Ground-R1 with its two ablated variants, Ground-R1-Intra and Ground-R1-Inter, which retain only the intra-bin or inter-bin reward, respectively. As shown in Table \ref{tab:ablate}, both variants underperform the full Ground-R1 model across all benchmarks, indicating that neither component alone is sufficient. In particular, Ground-R1-Intra exhibits notable degradation, suggesting that intra-bin normalization alone cannot adequately distinguish evidence across different scales. 

\noindent \textbf{\emph{Q5:}} \emph{Compared to GRPO, does SRPO lead to more accurate grounded evidence region?}

To investigate this, we analyze the training dynamics of Ground-R1 trained with SRPO and standard GRPO. As shown in Figure \ref{fig:visCurve}, we report the normalized average reward, grounding IoU, and tool call count over the course of training. Following \cite{zheng2025deepeyes}, grounding IoU is evaluated on the fine-grained visual search split of the training dataset, which provides ground-truth bounding boxes closely aligned with the target answers.

As shown in Figure \ref{fig:visCurve}, our SRPO consistently achieves higher average rewards and substantially improves grounding IoU compared to GRPO, indicating more accurate and faithful evidence region grounding. Notably, this performance gap becomes particularly pronounced after approximately 600 training steps. We attribute this to the inherent optimization bias of GRPO, which primarily favors large and visually salient bounding boxes (\cf Figure \ref{fig:regionReward}). As training progresses, the policy increasingly exploits previously high-reward large regions, leading to an \emph{exploration collapse} that prevents revisiting small but potentially critical evidence.


\subsection{Visualizations}  \label{sec:4.5}

Figure \ref{fig:inferCompare} provides a qualitative comparison between Ground-R1 trained with standard GRPO and our proposed SRPO. Under GRPO, Ground-R1 fails to localize the relevant evidence and instead grounds its reasoning on a large, visually salient region containing multiple houses and background context. As a result, the model defaults to an uncertain or incorrect response. In contrast, Ground-R1 trained with SRPO exhibits more precise evidence grounding by focusing on a small, task-relevant region where the vehicles are visible. By accurately localizing the silver vehicle and leveraging relative size cues, SRPO enables the model to correctly determine the relative distance to the camera.


\section{Conclusion}
In this work, we investigate a fundamental limitation of existing thinking-with-images approaches, namely a systematic scale-driven bias that favors large visual regions during optimization. To address this issue, we propose Ground-R1, equipped  with scale relative policy optimization, which recalibrates reward signals across evidence regions of different sizes. Extensive experiments demonstrate that Ground-R1 not only improves response accuracy but also yields more precise and faithful evidence grounding, offering a more robust pathway toward trustworthy and interpretable vision-language reasoning. 

\section*{Ethical Considerations} 
Our work contributes to enhancing the reliability and interpretability of LVLMs through a de-biased thinking-with-images pattern. By explicitly mitigating the scale-driven bias, Ground-R1 with SRPO reduces the model's tendency to rely on visually prominent yet semantically irrelevant regions, thereby lowering the risk of generating predictions based on spurious correlations. This advancement in evidence-based reasoning is particularly relevant for high-stakes applications, such as medical imaging analysis or autonomous systems, where overlooking subtle but critical details could have serious consequences. All data used in our experiments are from publicly available benchmark datasets intended for research purposes, and no personally identifiable information is involved. However, as with any foundation model enhancement, the improved capability could be deployed in dual-use scenarios. We advocate for responsible development and application of such technologies, including thorough testing in domain-specific contexts before real-world deployment. 

\section*{Limitation} 
While Ground-R1 significantly improves evidence grounding and response accuracy, our approach has the following two limitations that warrant further investigation. First, Ground-R1 remains dependent on the quality of region proposals, \ie, suboptimal proposals or missing candidate regions may still lead to incorrect grounding, particularly for highly occluded or visually ambiguous objects. Second, our evaluation focuses on established benchmarks while the performance and grounding fidelity of Ground-R1 in more open-world, long-tail, or domain-shifted scenarios remain to be explored. Future work should aim to integrate more robust region generation strategies and extend evaluations to broader real-world settings with greater visual and semantic diversity.
\bibliography{refs.bib}
\clearpage
\appendix

\section{Appendix}

This appendix provides additional experimental analyses and qualitative visualizations to complement the main paper: \begin{itemize}[topsep=0pt, partopsep=0pt, leftmargin=20pt, parsep=0pt, itemsep=3pt]
    \item Hyper-parameter Analysis.
    \item Prompts for Ground-R1.
    \item Illustrations of the intra-bin reward design.
    \item Visualizations of completion length.
    \item Qualitative comparisons.
\end{itemize}

\noindent\textbf{Hyper-parameter analysis.} Table \ref{tab:ablateParam} presents an ablation study on the key hyper-parameters of Ground-R1, including the number of grounding rollouts $G_1$, answer rollouts $G_2$, and the number of scale bins $K$. We observe that a balanced allocation between grounding and answering stages is crucial for optimal performance. In particular, the configuration $G_1{=}4$ and $G_2{=}2$ achieves the best overall results across all high-resolution benchmarks. Moreover, varying the number of scale bins shows that $K{=}3$ consistently outperforms smaller or larger values, suggesting that moderate scale discretization provides a favorable balance between granularity and training stability. 

\noindent \textbf{Prompts for Ground-R1}. The prompts for Ground-R1 is illustrated in Table \ref{tab:Prompt}.

\noindent \textbf{Illustrations of the intra-bin reward design.} The intra-bin reward in SRPO  integrates two complementary reward mechanisms: 
\begin{itemize}[topsep=0pt, partopsep=0pt, leftmargin=10pt, parsep=0pt, itemsep=3pt]
    \item \emph{The format reward} ensures coherence by structuring reasoning traces within \think{and} tags, spatial coordinates within \bbox{and} tags, and final answers within \answer{and} tags. 

    \item \emph{The answer reward} is defined according to the question type: for multiple-choice questions, it is a binary score based on exact answer matching, while for free-form questions, it is computed as lexical alignment using the average of ROUGE-1, ROUGE-2, and ROUGE-L scores against the ground-truth answers.
\end{itemize}

\begin{figure}[t]
        \centering
         \includegraphics[width=0.49\textwidth]{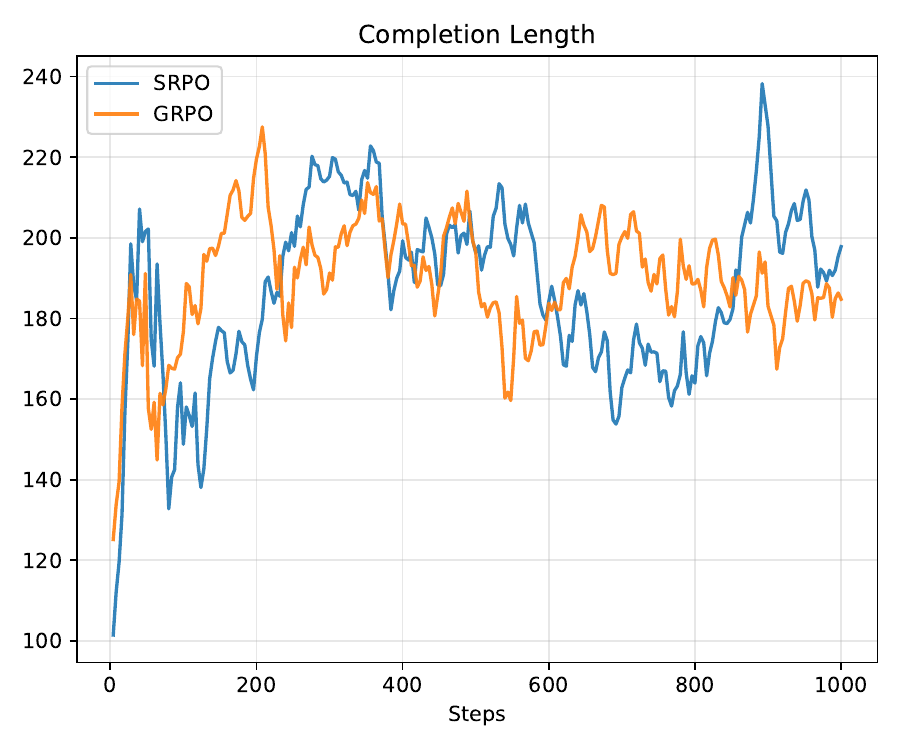}
        \caption{Training dynamics of completion length for Ground-R1 under SRPO and standard GRPO.}
        \label{fig:visCompletion}
\end{figure}

\noindent \textbf{Visualizations of completion length.} Figure \ref{fig:visCompletion} depicts the training dynamics of completion length for Ground-R1 trained with SRPO and standard GRPO. As shown, both methods exhibit comparable completion length trajectories throughout training, with similar magnitudes and fluctuations across steps. This observation indicates that the performance improvements achieved by SRPO are not attributable to longer or more verbose generations. Instead, SRPO enhances reasoning quality and evidence grounding without increasing output length or introducing additional inference overhead.

\begin{table}[t]
\belowrulesep=0pt
\aboverulesep=0pt
    \centering
    \caption{Ablations of hyper-parameters.}
    \setlength{\tabcolsep}{4mm}
    \resizebox{0.49\textwidth}{!}{
    \begin{tabular}{ccc|ccccccc}
    \toprule
    \rowcolor{gray!10} 
    \textbf{$G_1$}  & \textbf{$G_2$} & \textbf{$K$} & \textbf{V$^{*}$} & \textbf{HR-4K}  & \textbf{HR-8K}\\
    \hline
    1 & 8 & 3 & 77.0 & 67.0 & 59.1 \\
    2 & 4 & 3 & 83.8 & 73.8 & 69.5\\ 
    \textbf{4} & \textbf{2} & \textbf{3} & \textbf{87.4} & \textbf{75.0} & \textbf{71.1} \\
    8 & 1 & 3 & 86.4 & 74.7 & 70.6\\
    4 & 2 & 2 & 86.9 & 73.5 & 70.0\\
    4 & 2 & 4 & 86.9 & 74.1 & 70.8\\
    \bottomrule
    \end{tabular}}
    \label{tab:ablateParam}
\end{table}

\begin{table*}[ht]
\centering
\caption{\textbf{Prompts for Ground-R1.} \textcolor{red}{\{input\}} will be replaced with the specific question and image during training and inference.} 
        \centering
        \resizebox{0.95\textwidth}{!}{
        \begin{tabular}{p{15cm}}
        \hline
        Question: \textcolor{red}{\{input\}} \\
        Please think about this question as if you were a human pondering deeply. Provide detailed reasoning between the \think{and} tags. Reason whether there exists a region in the image that could help you answer the question better. If such a region exists, provide one bounding box coordinate in the format [x1,y1,x2,y2] inside the \bbox{and} tags. \\
        Then, you will receive a cropped image based on the bounding box. Use all the images to continue reasoning. You may conduct multiple rounds of grounding to refine your region as you want. The bounding box you provide should always be selected based on the original image. If at any point you determine no further visual information is needed, you may directly provide the final answer inside the \answer{and} tags.\\
        Format Example:\\
        \think{Reasoning Process} \bbox{[x1,y1,x2,y2]} or, \\
        \think{Reasoning Process}\answer{Final Answer} \\ 
        \hline
        \end{tabular}
        }
     \label{tab:Prompt}
\end{table*}


\begin{figure*}[ht]
	\centering
        \includegraphics[width=0.9\textwidth]{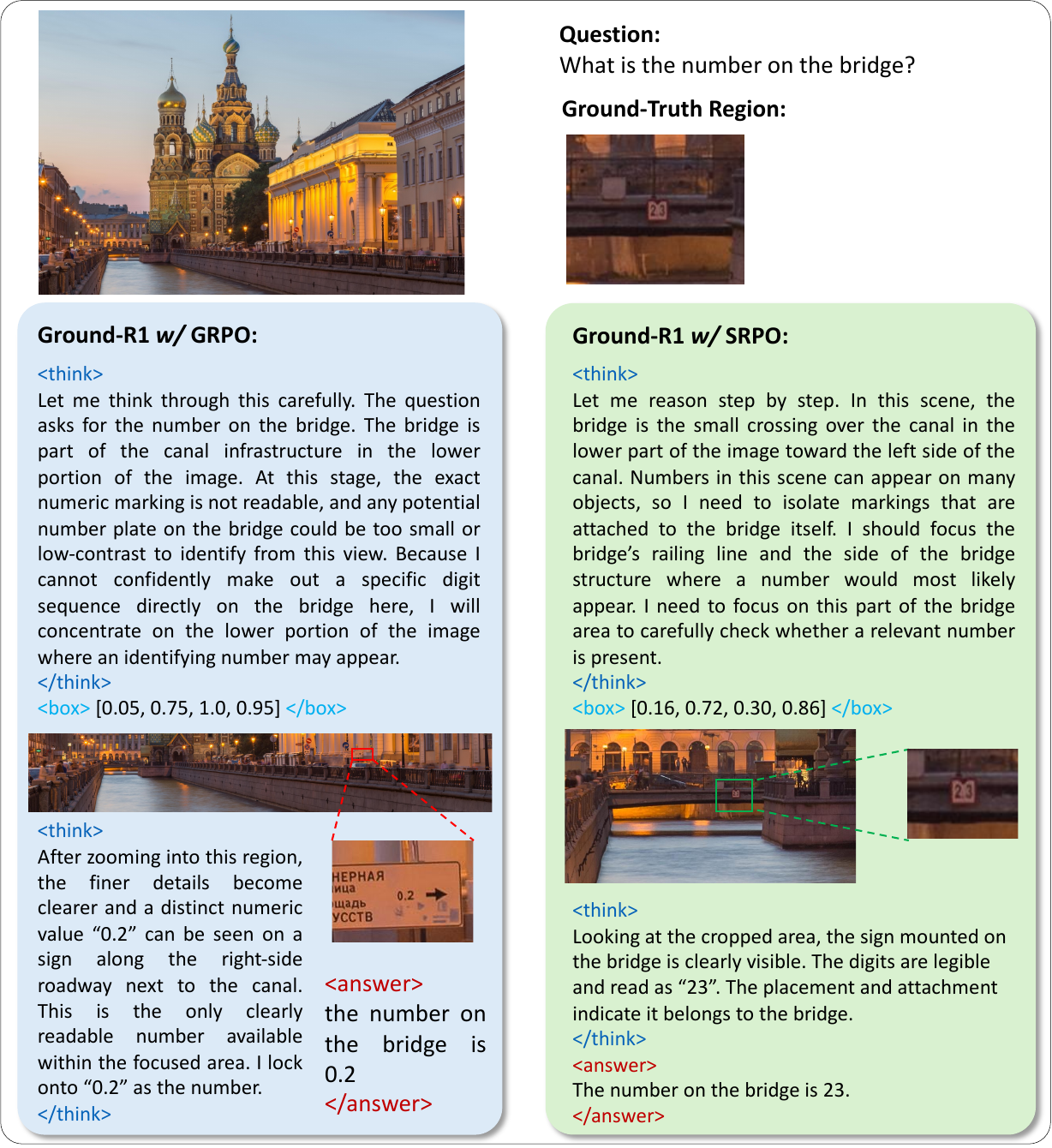}
	\caption{\textbf{Comparisons of reasoning trajectories} of Ground-R1 under standard GRPO and our proposed SRPO.}
	\label{fig:inferCompareAppendix2}
\end{figure*}

\begin{figure*}[ht]
	\centering
        \includegraphics[width=0.9\textwidth]{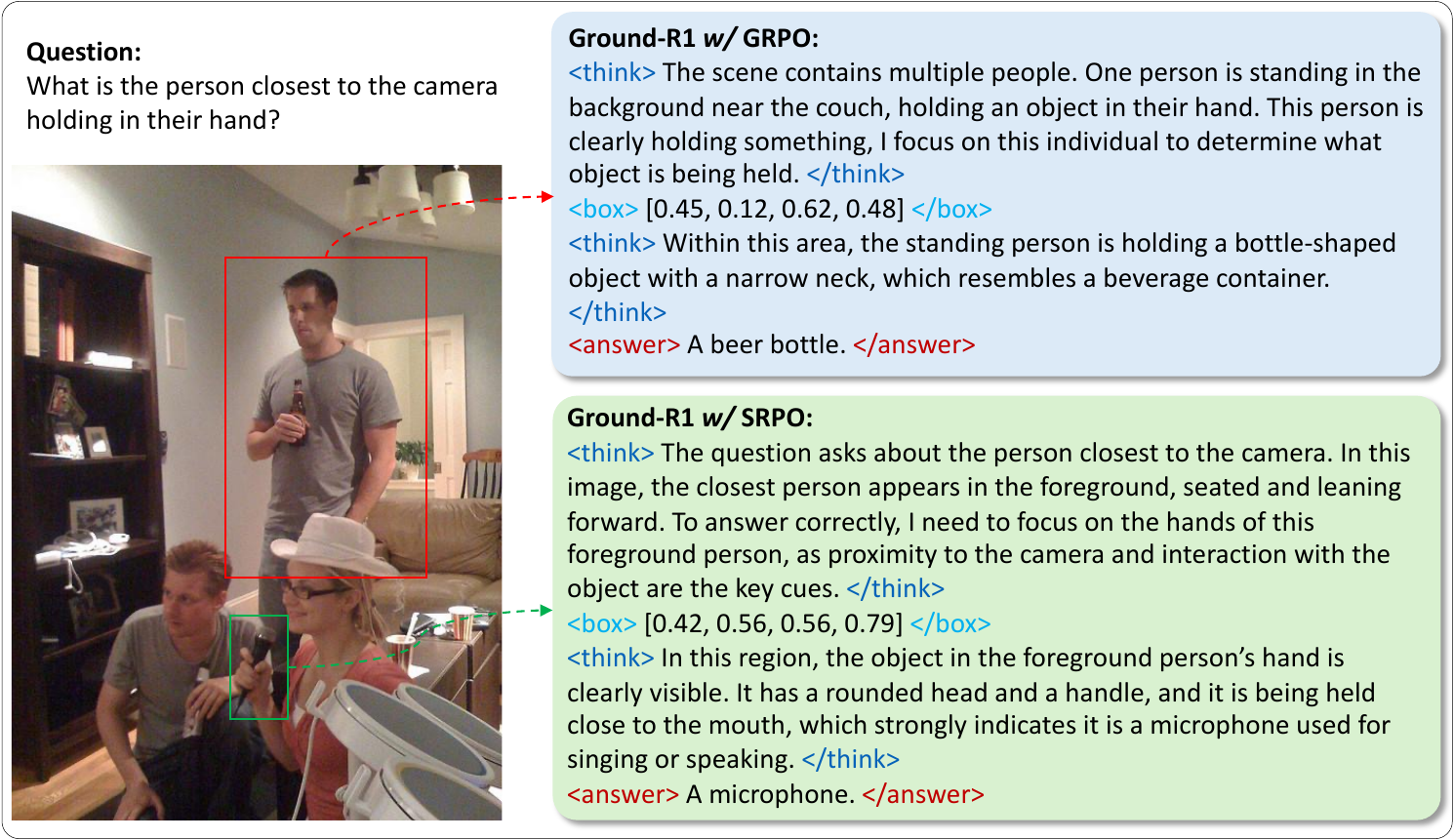}
	\caption{\textbf{Comparisons of reasoning trajectories} of Ground-R1 under standard GRPO and our proposed SRPO.}
	\label{fig:inferCompareAppendix3}
\end{figure*}

\begin{figure*}[ht]
	\centering
        \includegraphics[width=0.9\textwidth]{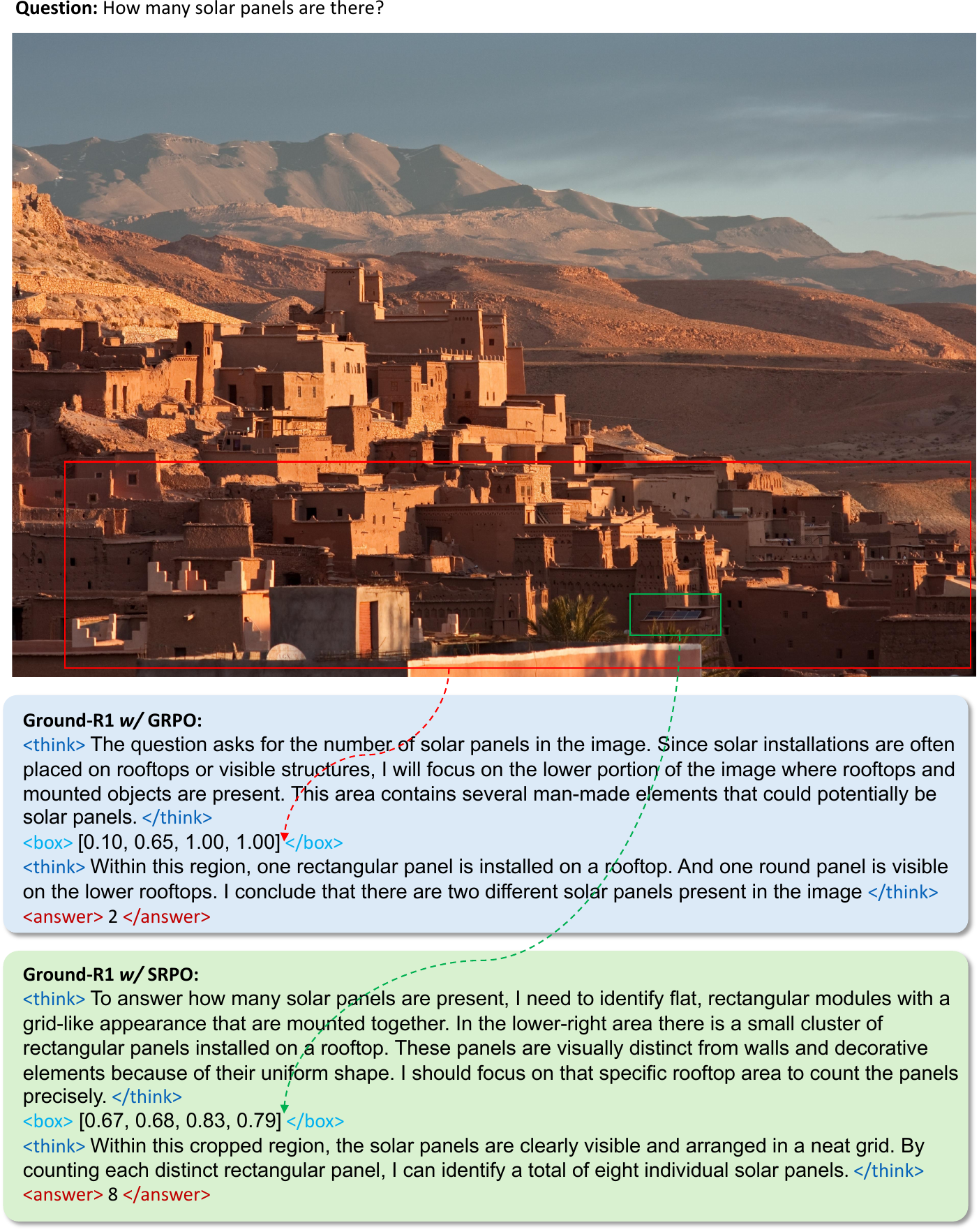}
	\caption{\textbf{Comparisons of reasoning trajectories} of Ground-R1 under standard GRPO and our proposed SRPO.}
	\label{fig:inferCompareAppendix4}
\end{figure*}

\noindent \textbf{Qualitative comparisons.} Figures \ref{fig:inferCompareAppendix2}–\ref{fig:inferCompareAppendix4} present qualitative comparisons of reasoning trajectories produced by Ground-R1 trained with standard GRPO and our proposed SRPO. Figure \ref{fig:inferCompareAppendix2} focuses on a fine-grained recognition scenario that requires identifying a small numerical marker on a bridge. Under standard GRPO, the model grounds its reasoning on a large, visually salient region and incorrectly associates an unrelated numeric sign with the bridge. In contrast, SRPO enables more precise localization of the relevant bridge region, allowing the model to correctly identify the target number. Figure \ref{fig:inferCompareAppendix3} illustrates a foreground object identification task involving multiple people in the scene. While GRPO grounds on a prominent background figure and produces an incorrect answer, SRPO successfully focuses on the closest person to the camera and correctly recognizes the object being held. Figure \ref{fig:inferCompareAppendix4} further examines a counting task involving small-scale objects, where GRPO overgeneralizes from a large region and undercounts the solar panels, whereas SRPO isolates the small but relevant rooftop region and accurately counts all instances.

Overall, these examples consistently demonstrate that SRPO mitigates the tendency of standard GRPO to favor large, visually salient regions and instead encourages grounding on small yet semantically critical evidence. As a result, Ground-R1 trained with SRPO exhibits more faithful evidence localization and more reliable reasoning outcomes across diverse visual reasoning scenarios.




\appendix

\end{document}